\documentclass[twoside,leqno,twocolumn]{article}

\usepackage[letterpaper]{geometry}

\usepackage{ltexpprt}
\usepackage{graphicx}

\begin{document}

\title{\Large A Framework for Unsupervised Classification and Data Mining of Tweets about Cyber Vulnerabilities}
\author{Kenneth Alperin\thanks{MIT Lincoln Laboratory, kenneth.alperin@ll.mit.edu}
\and Emily Joback\thanks{MIT Lincoln Laboratory}
\and Leslie Shing\thanks{MIT Lincoln Laboratory}
\and Gabe Elkin\thanks{MIT Lincoln Laboratory}}

\date{}

\maketitle


\fancyfoot[R]{\scriptsize{Copyright \textcopyright\ 2021 by SIAM\\
Unauthorized reproduction of this article is prohibited}}





\begin{abstract} \small\baselineskip=9pt
Many cyber network defense tools rely on the National Vulnerability Database (NVD) to provide timely information on known vulnerabilities that exist within systems on a given network. However, recent studies have indicated that the NVD is not always up to date, with known vulnerabilities being discussed publicly on social media platforms, like Twitter and Reddit, months before they are published to the NVD. To that end, we present a framework for unsupervised classification to filter tweets for relevance to cyber security. We consider and evaluate two unsupervised machine learning techniques for inclusion in our framework, and show that zero-shot classification using a Bidirectional and Auto-Regressive Transformers (BART) model outperforms the other technique with 83.52\% accuracy and a F1 score of 83.88, allowing for accurate filtering of tweets without human intervention or labelled data for training. Additionally, we discuss different insights that can be derived from these cyber-relevant tweets, such as trending topics of tweets and the counts of Twitter mentions for Common Vulnerabilities and Exposures (CVEs), that can be used in an alert or report to augment current NVD-based risk assessment tools.
\end{abstract}

\section{Introduction}

The NVD publishes open source information about cyber vulnerabilities as they are reported, and many network defense tools use the NVD as a data source to update organizations of risks and threats to a given network. For example, if a network contains computers that use Windows operating systems, a tool might reference the NVD to alert users on known weaknesses in Windows that a hacker could potentially use as a foothold to access the network. Such alerts will inform network security analysts to patch the system before a malicious actor can exploit the vulnerability. Therefore, it is critical that the NVD is updated as soon as a vulnerability is discovered in order to give security analysts an advantage against cyber attackers. However, research shows the NVD is not always updated in a timely manner, and that information about vulnerabilities is often openly discussed on social media platforms months before it is published by the NVD \cite{sabottke2015vulnerability, sauerwein2018tweet, alves2020follow}. These social media platforms, such as Twitter, Reddit, and GitHub, have become profilic sources of information due to the wide access and reach of these mediums \cite{shrestha2020multiple, horawalavithana2019mentions}. These platforms can be leveraged as open source threat intelligence (OSINT) for use by cyber operators.

Vulnerabilities are given a Common Vulnerabilities and Exposures identifier (CVE ID) when entered into the NVD. However, social media platforms, like Twitter, may often exclude the CVE ID in their discussions of vulnerabilities that have not yet been published to the NVD. Even after a vulnerability is in the NVD, related discussions on Twitter may still lack the CVE ID. Therefore, the data gathered for analysis of tweets with trending vulnerability discussions should not be limited to tweets that contain CVE IDs, but encompass all tweets related to cyber vulnerabilities. To address this, we present a framework for unsupervised relevance filtering of `vulnerability' tweets and insights that can be gained from mining cyber-relevant tweets. The contributions of our work include:
\begin{itemize}
    \item An implementation of two unsupervised methods evaluated on a labelled `vulnerability' Twitter dataset, with the zero-shot BART classifier yielding an 83.52\% accuracy. To our knowledge, prior machine learning (ML) techniques for cyber-relevance filtering require labelled data for supervised methods or fine-tuning on a cyber-relevant dataset; 
    \item Unique insights that can be gained from cyber-relevant tweets, including topic analysis and counts of tweets per CVE, and a use case showing how tweet counts can prioritize CVEs with low Common Vulnerability Scoring System (CVSS3) \cite{cvss_v3_1_specification_document} scores.
\end{itemize}
Together, this relevance filtering and data mining framework allows for real-time situation awareness (SA) of vulnerability information from Twitter that can supplement existing tooling. Finally, we discuss ways to extend this work into operational use.

We first provide an overview of prior work related to our analysis in Section 2. In Section 3, we discuss our approach to implementing and evaluating different unsupervised ML methods for relevance filtering. In Section 4, we discuss insights gained from mining cyber-relevant tweets with examples. In Section 5, we discuss our findings and limitations, and highlight future research directions to build on this work. 

\section{Related Work}

Prior work exists in applying ML to filter for cyber-relevant tweets. To our knowledge, they all require labelled data for training in supervised methods or fine-tuning on a cyber-relevant dataset. Supervised classification has been used on tweets related to IT assets \cite{dionisio2019cyberthreat}, tweets containing at least one security keyword from certain user accounts \cite{alves2021processing}, tweets from cyber experts with n-grams extracted \cite{queiroz2017predicting}, and social media threats from multiple platforms \cite{lippmann2016finding}.
Mittal et al. use a trained named entity recognizer to extract security terms from tweets, and then label tweets with at least two security terms as cyber-relevant \cite{mittal2016cybertwitter}. On the labelled dataset we use, Behzadan et al. achieve 94.72\% accuracy \cite{behzadan2018corpus} and Simran et al. get 89.3\% with word embeddings \cite{simran2019deep}, but these methods requires training and labelled data. 
Shin et al. use word embeddings and fine-tune with cyber security positive and negative data, but do not specifically focus on tweets \cite{shin2020new}. Iorga et al. classified news articles as cyber-relevant using a fine-tuned BERT model \cite{9324852}. 

There is also prior work in mining features for analysis from cyber-relevant tweets. Rodriguez et al. use keywords to filter for cyber-relevance and active learning to update the keywords, and use tweet sentiment as a feature over time \cite{rodriguez2020social}. Zong et al. study the connection of severity of cyber threats with how they are discussed online \cite{zong2019analyzing}. Hoppa et al. visualize cyber-relevant tweets in Kibana for end users \cite{hoppa_debb_hsieh_kc_2019}. Alves et al. cluster cyber-relevant tweets, select exemplar tweets for each cluster, enrich those tweets with extracted CVEs, associated CVSS impact scoring and publish date, \cite{alves2021processing}, and count Twitter mentions per CVE \cite{alves2020follow}. There is also research in generating alerts and indicators of compromise (IoCs) from tweets \cite{mittal2016cybertwitter, dionisio2019cyberthreat}. Sabottke et al. analyze exploits associated with vulnerabilities from Twitter \cite{sabottke2015vulnerability}.

\section{Relevance Filtering} 

The goal of relevance filtering is to reduce the large corpus of topically-varied tweets down to a more manageable and cyber-relevant set for downstream data mining and analysis. In this section, we describe two unsupervised ML methods considered for inclusion in our pipeline, evaluate their performances in correctly classifying tweets as cyber-relevant or not using a labelled dataset, and discuss tradeoffs of the methods.

\subsection{Classification Techniques}
We considered two unsupervised classification techniques for relevance filtering. We focused exclusively on unsupervised methods for a couple reasons. First, we wanted to develop a filtering pipeline that could be used in a real-time operational setting and be resilient to the addition of new data points. The performance of supervised ML models tends to degrade over time when the distribution of new data points differs widely from that of the model's training data. In particular, Twitter topics can trend in varying degrees over different time periods, so it is important to have a robust classification model that can account for this variation. Second, supervised ML models require labelled data to train the models. These datasets are not readily available and labelling is a time-intensive and manual process. Thus, to overcome these short-comings, we investigated the inclusion of k-means clustering and zero-shot classification into our pipeline. 

k-means clustering is a widely used technique for identifying subgroups of "similar" points within a larger dataset \cite{hartigan1979ak}. The number of clusters, k, is set by the user. The algorithm then determines the groupings by identifying \textit{k} cluster centroids such that the sum of the squared Euclidean distance between points within each cluster and the corresponding centroid is minimized. For this method, we first used Term Frequency-Inverse Document Frequency (TF-IDF) to convert each tweet to a vector representation for the numeric importance of each term to each document \cite{ramos2003using}. Then those vectors were inputted into the k-means algorithm. To identify a cluster as cyber-relevant, we assumed there must be at least one tweet in the cluster with a CVE mentioned. 

One of the more recent advances in language modeling is zero-shot classification. This method applies language models, such as BART \cite{lewis2019bart}, to associate input text with a user-specified hypothesis, such as `This text is related to cyber security' (the prompt used for our analysis), and generates a likelihood of the hypothesis being correct. In this way, the language model acts as a classifier and does not require any training by the user since the model comes pre-trained on large amounts of sentence/hypothesis pairs. Specifically, we tested two different BART models. The bart-large-mnli model comes pre-trained on the Multi-Genre Natural Language Inference (MultiNLI) corpus \cite{N18-1101}, which contains 433,000 sentences paired with hypotheses. The distilbart-mnli-12-3 model is distilled from the bart-large-mnli model, containing fewer parameters and runs faster, but is slightly less accurate.


\subsection{Dataset}

In this experiment, the unsupervised techniques were evaluated on a labelled dataset of tweets from Behzadan et al \cite{behzadan2018corpus}. The dataset contains 21,368 tweets  collected over four days using common cyber security keywords, and were labelled as `threat', `business', `unknown', and `irrelevant'. Since we focus on vulnerabilities, we first filtered the dataset for tweets that contain the term `vulnerability', which came out to 9,963 tweets. Tweets labelled as `business', `unknown' and  `threat' were replaced with a `cyber-relevant' label as they also appeared to be relevant to cyber security, and comprised 54.5\% of the filtered dataset.

\subsection{Results}

\begin{table}[h!]
  \vspace{-8mm}
  \begin{center}
    \caption{Relevance Filtering Classification Results}
    \label{tab:results}
    \begin{tabular}{c|c|c|c} 
      \textbf{Method} & \textbf{Parameter} & \textbf{Acc.} & \textbf{F1}\\ 
      \hline
      k-means & k=2 & 55.21 & 70.74\\
      k-means & k=5 & 54.01 & 70.06\\
      k-means & k=10 & 53.46 & 68.62\\
      Zero-Shot & bart-large-mnli & 83.52 & 83.88\\
      Zero-Shot & distilbart-mnli-12-3 & 83.40 & 83.53\\
    \end{tabular}
  \end{center}
  \vspace{-8mm}
\end{table}

Table \ref{tab:results} shows the accuracy and F1 score for each method evaluated against the labelled dataset. For the k-means pipeline, we evaluated values 2, 5, and 10 for k, and used the previously described method for labelling clusters as cyber-relevant. All the methods are over 99\% accurate in classifying tweets with explicit CVE mentions as cyber-relevant, which comprises 12.49\% of the dataset. However, on all `vulnerability' tweets, the zero-shot classification models significantly outperform the k-means clustering approach, with the bart-large-mnli model resulting in the highest accuracy of 83.52\% and the highest F1 score of 83.53. On the set of tweets with no CVE mention, the bart-large-mlni model is 81.29\% accurate with a F1 score of 78.85. On a Volta GPU, the distilbart-mnli-12-3 takes 2.18 minutes to classify the dataset, whereas the bart-large-mnli model takes 3.64 minutes. Overall, the BART model can accurately predict cyber-relevance without any fine-tuning or supervised training, making it useful in a framework for classifying tweets in real-time. 

\section{Data Mining of Cyber-Relevant Tweets} 

With the ability to filter tweets for cyber-relevance in real-time using an unsupervised ML method, the next step is to mine the corpus for insights that could be useful for a cyber operator. In this section, we discuss the use of word embeddings to identify key topics and trends in our collected Twitter dataset, and the metric developed to identify the most discussed vulnerabilities based on CVE mention counts.

\subsection{Dataset}

\begin{figure}[tb]
  \centering
  \includegraphics[width=\linewidth]{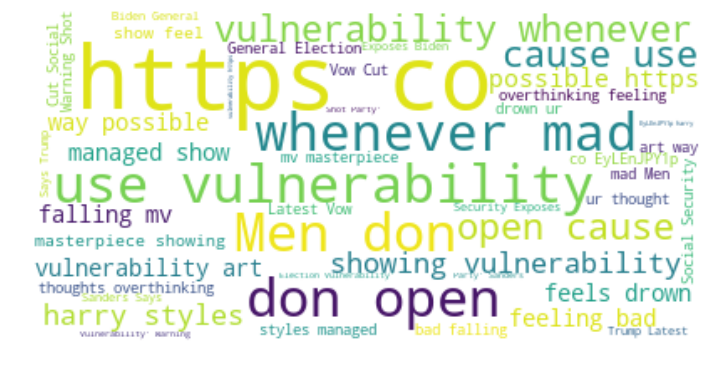}
  \caption{Most common phrases in tweets containing the term `vulnerability' from 2/19/20$-$3/18/20}
  \label{f:tweet_wc}
\end{figure}

To create a new dataset for analysis, we used the Twitter API to collect tweets in real-time that contained the keyword `vulnerability', which we hypothesized was a suitable term for producing tweets related to cyber vulnerabilities in our initial data collection. These tweets were collected from 2/19/20$-$3/18/20, and statistics for these are in Table \ref{tab:stats}. Figure \ref{f:tweet_wc} is a word cloud of the most common phrases across this corpus. The zero-shot bart-large-mnli classifier filtered out 76.83\% of tweets, producing 77,979 tweets classified as cyber-relevant for analysis. 
\begin{table}[h!]
  \vspace{-4mm}
  \begin{center}
    \caption{Characteristics of Collected Tweets}
    \label{tab:stats}
    \begin{tabular}{c|c} 
      \textbf{Statistic} & \textbf{Value}\\
      \hline
      Number of Tweets & 337,468\\
      Avg. Tweets per Day & 12,052\\
      Avg. Words per Tweet & 29\\
      \% English Tweets & 92\%\\
      \% Tweets with URL & 45\%\\
    \end{tabular}
  \end{center}
  \vspace{-8mm}
\end{table}
\subsection{Topic Analysis}

Word embeddings were used as an alternative to TF-IDF to numerically represent the text. While TF-IDF is useful at capturing a few key terms describing a given text, 
it does not account for words based on their relationships or contexts. For example, the terms `vulnerability' and 'vector' may seem unrelated when observed as a bag-of-words by TF-IDF but in context are actually related to a cyber attack surface. A word embedding model, like Word2Vec \cite{mikolov2013efficient}, detects this relationship and meaning by considering the surrounding words. Word2Vec is a neural network that learns a vector of weights for each term in a text corpus. Given two different words contained in nearly identical phrases, the weights learned for that word will be similar, and thus the two words will be close to each other in the vector space representation. The terms in each tweet were inputted to Word2Vec without any pre-processing, which created a word embedding. Clustering can then be applied to the word embeddings to yield topics within the full set of text data.

\begin{figure}[tb]
  \centering
  \includegraphics[width=\linewidth]{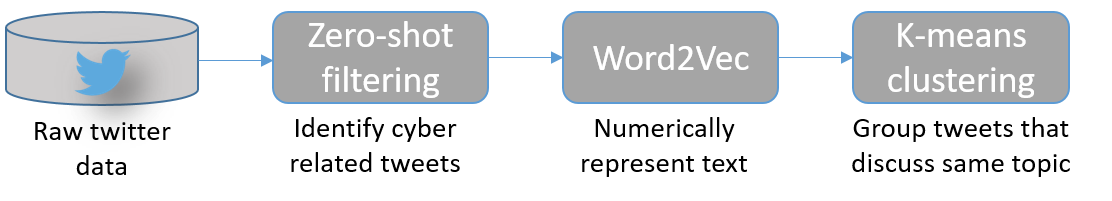} \caption{Topic Analysis Pipeline}
  \label{f:word2vec}
  \vspace{-4mm}
\end{figure}



Figure \ref{f:word2vec} shows our topic analysis pipeline. We used the elbow method in plotting the sum of squared errors (SSE) for different k values to select the optimal k, 
which ended up being 10. 
There are 31,821 unique tweets in this corpus, but clustering only those showed little difference.
Table \ref{tab:clusters} shows three of the clusters generated with the number of tweets and most common key terms. In particular, we see a cluster centered around the KR00K CVE \cite{krook}, and two clusters for different types of Microsoft vulnerabilities. These initial clusters show promise in topic identification, but further analysis is warranted to determine optimal clustering parameters. These clusters allow for quick triaging of vulnerability information and enable an analyst to efficiently focus on certain topics. 

\begin{table}[h!]
  \vspace{-4mm}
  \begin{center}
    \caption{Topic Analysis Keywords for Example Clusters}
    \label{tab:clusters}
    \begin{tabular}{c|c|c} 
      \textbf{ID} & \textbf{Total Tweets} & \textbf{Keywords}\\
      \hline
      1 & 8,701 & patch, Microsoft, SMBv3\\
      5 & 4,414 & devices, Wi-Fi, KR00K\\
      7 & 4,995 & remote, code, Microsoft\\
    \end{tabular}
  \end{center}
  \vspace{-6mm}
\end{table}
\subsection{CVE Tweet Counts}

In addition to trending topics, we also implemented metrics to determine the degree of importance of certain vulnerability mentions extracted from our curated cyber-relevant Twitter corpus. In particular the `tweet count` for a CVE observes the number of times a CVE is mentioned in the tweets across a specified time period. The tweet count can indicate public interest in the vulnerability, and serve as a potential proxy to measuring the exploitability of the vulnerability. For instance, if many users are discussing a CVE, it is possibly an active threat on their networks. 

\begin{table}[h!]
  \begin{center}
    \vspace{-4mm}
    \caption{Most Mentioned CVEs on Collected Tweets}
    \label{tab:table1}
    \begin{tabular}{c|c|c} 
      \textbf{CVE ID} & \textbf{CVSS3} & \textbf{Tweet Count}\\
      \hline
      CVE-2020-0688 & 8.8 & 1621\\
      CVE-2020-1938 & N/A & 1014\\
      CVE-2020-8597 & 9.8 & 741\\
      CVE-2020-8794 & 9.8 & 349\\
      CVE-2019-15126 & 3.1 & 279\\
    \end{tabular}
  \end{center}
  \vspace{-4mm}
\end{table}

Many vulnerability management tools rely on CVSS3 scoring to prioritize vulnerabilities on a network, since cyber defenders do not have time and resources to patch every vulnerability. Table \ref{tab:table1} shows the five most mentioned CVEs from the tweets we collected, along with their CVSS3 scores and tweet counts during the time period. We also calculated the correlation of the tweet counts and the CVSS3 scores for the CVEs in our corpus, which came out to .04, indicating no correlation. The fifth most mentioned vulnerability, CVE-2019-15126, is called the KR00K CVE, and allows unauthorized decryption in WiFi chips. Even though it has a low CVSS3 score of 3.1, the CVE has affected over 1 billion devices from companies such as Amazon, Apple, and Google, and has a known exploit \cite{krook}. The tweet count metric can detect exploited CVEs otherwise not prioritized by traditional risk management metrics. 

\section{Discussion and Future Work} 

The zero-shot classification pipeline was able to accurately filter out a large percentage of tweets and retain the cyber-relevant samples, without any labelled training data. Future work on the pipeline includes expanding the Twitter data collection to include other key terms, such as `exploit', and expand to data collected from Reddit. Tweets about cyber vulnerabilities can be mined for useful insights, such as the tweet count per CVE. This metric can detect some highly exploited CVEs otherwise not prioritized by other risk management metrics. Future work includes identifying other attributes about authors of tweets, such as their country of origin and association with security organizations, to weight the validity and importance of collected tweets, and analyzing the difference in chatter around vulnerabilities before and after they are published in the NVD.  

We plan to incorporate these metrics and analyses into a reporting dashboard or alerting tool that operators can use in real-time for improved SA. The information from tweets can augment current risk assessment tools that rely on the NVD, since social media enrichment reporting provides analysts additional CVE information and additional patch actions not possible before through links shared in the posts. Future work includes developing a user interface to generate reports.

\section{Conclusion} 

In this paper, we show how unsupervised zero-shot classification using the BART language model allows for accurate prediction of cyber-relevance for tweets containing the term `vulnerability', with an accuracy of 83.52\%. This is significant since this method does not require any manual labelling of data for training or fine-tuning of a model. Cyber-relevant tweets can be mined to extract information and insights that are useful in cyber SA. One such insight is the tweet count metric, and we show how it prioritizes vulnerabilities that otherwise would not be highlighted using traditional ranking methods such as CVSS3. Social media data can augment current risk management tooling, and we plan to incorporate our findings from this paper into a reporting dashboard or alerting tool that would provide cyber operators with real-time SA of information disseminated on social media platforms.

\section{Acknowledgments}
DISTRIBUTION STATEMENT A. Approved for public release. Distribution is unlimited. This material is based upon work supported by the Federal Aviation Administration under Air Force Contract No. FA8702-15-D-0001. Any opinions, findings, conclusions or recommendations expressed in this material are those of the author(s) and do not necessarily reflect the views of the Federal Aviation Administration. Delivered to the U.S. Government with Unlimited Rights, as defined in DFARS Part 252.227-7013 or 7014 (Feb 2014). Notwithstanding any copyright notice, U.S. Government rights in this work are defined by DFARS 252.227-7013 or DFARS 252.227-7014 as detailed above. Use of this work other than as specifically authorized by the U.S. Government may violate any copyrights that exist in this work.


\end{document}